\pdfoutput=1

\PassOptionsToPackage{prologue,dvipsnames}{xcolor}

\documentclass[11pt]{article}

\usepackage[final]{acl}

\usepackage{times}
\usepackage{latexsym}

\usepackage[T1]{fontenc}

\usepackage[utf8]{inputenc}

\usepackage{microtype}

\usepackage{inconsolata}

\usepackage{graphicx}
\usepackage{dsfont}

\usepackage{enumitem}
\usepackage{rotating}

\usepackage{booktabs}
\usepackage{multirow}
\usepackage{subcaption}

\usepackage{arydshln}

\usepackage{amsthm}
\usepackage{amsmath}
\usepackage{amssymb}
\usepackage{xspace}
\newcommand{\narrowtextsc}[1]{\textls[-50]{\textsc{#1}}}
\newcommand{\lm}[1]{\texttt{#1}}
\newcommand{\sys}[1]{\narrowtextsc{#1}}
\newcommand{\data}[1]{\textsf{#1}}

\usepackage{colortbl}
\definecolor{Gray}{gray}{0.94}
\definecolor{LightCyan}{rgb}{0.88,1,1}
\newcolumntype{a}{>{\columncolor{Gray}}c}
\newcolumntype{o}{>{\columncolor{white}}c}
\usepackage{soul}
\definecolor{celeste}{cmyk}{0.3922, 0.0353, 0, 0.1}
\definecolor{purple}{cmyk}{0.16, 0.28, 0, 0}
\definecolor{brilliantlavender}{cmyk}{0, 0.2235, 0, 0.1}
\definecolor{LightRed}{RGB}{232, 56, 107} 
\definecolor{LightBlue}{RGB}{116, 232, 226}
\definecolor{Tan}{rgb}{0.8203,0.7031,0.5469}
\definecolor{gblue}{RGB}{81,231,195}

\definecolor{greenblue}{RGB}{142, 207,201}

\definecolor{orange}{RGB}{255, 190, 122}

\definecolor{red}{RGB}{250, 127,111}

\definecolor{blue}{RGB}{130, 176, 210}

\usepackage{verbatim}

\title{\sys{FitCF}: A Framework for Automatic Feature Importance-guided Counterfactual Example Generation}

\newcommand{\affilsup}[1]{\rlap{\textsuperscript{\normalfont#1}}}

\author{
    Qianli Wang\affilsup{1,2,\footnotemark[2]}
    \qquad 
    Nils Feldhus\affilsup{1,2,6}
    \qquad
    Simon Ostermann\affilsup{2,4,5}
    \\
    \textbf{Luis Felipe Villa-Arenas\affilsup{1,2,3}}
     \qquad
    \textbf{Sebastian M\"oller\affilsup{1,2}}
    \qquad
    \textbf{Vera Schmitt\affilsup{1,2}}
    \\
    $^1$Quality and Usability Lab, Technische Universit\"at Berlin \\
    $^2$German Research Center for Artificial Intelligence (DFKI) 
    \quad
    $^3$Deutsche Telekom\\
    $^4$Saarland Informatics Campus
    \quad
    $^5$Centre for European Research in Trusted AI (CERTAIN) \\
    $^6$BIFOLD – Berlin Institute for the Foundations of Learning and Data\\
  \small{\textbf{\footnotemark[2] Correspondence}: 
  \texttt{\href{mailto:qianli.wang@tu-berlin.de}{qianli.wang@tu-berlin.de}}}
}

\begin{document}
\maketitle
\begin{abstract}
Counterfactual examples are widely used in natural language processing (NLP) as valuable data to improve models, and in explainable artificial intelligence (XAI) to understand model behavior. The automated generation of counterfactual examples remains a challenging task even for large language models (LLMs), despite their impressive performance on many tasks. 
In this paper, we first introduce \sys{ZeroCF}, a faithful approach for leveraging important words derived from feature attribution methods to generate counterfactual examples in a zero-shot setting. Second, we present a new framework, \sys{FitCF}\footnote{Code is available at: \url{https://github.com/qiaw99/FitCF}. \sys{FitCF} stands for ``\underline{F}eature \underline{I}mpor\underline{t}ance-guided \underline{C}ounter\underline{f}actual Example Generation''.}, which further verifies aforementioned counterfactuals by label flip verification and then inserts them as demonstrations for few-shot prompting, outperforming three state-of-the-art baselines. Through ablation studies, we identify the importance of each of \sys{FitCF}'s core components in improving the quality of counterfactuals, as assessed through flip rate, perplexity, and similarity measures. Furthermore, we show the effectiveness of \textit{LIME} and \textit{Integrated Gradients} as backbone attribution methods for \sys{FitCF} and find that the number of demonstrations has the largest effect on performance. Finally, we reveal a strong correlation between the faithfulness of feature attribution scores and the quality of generated counterfactuals, which we hope will serve as an important finding for future research in this direction.


\end{abstract}

\begin{figure}[t!]
\centering
\resizebox{\columnwidth}{!}{
\begin{minipage}{\columnwidth}
\includegraphics[width=\columnwidth]{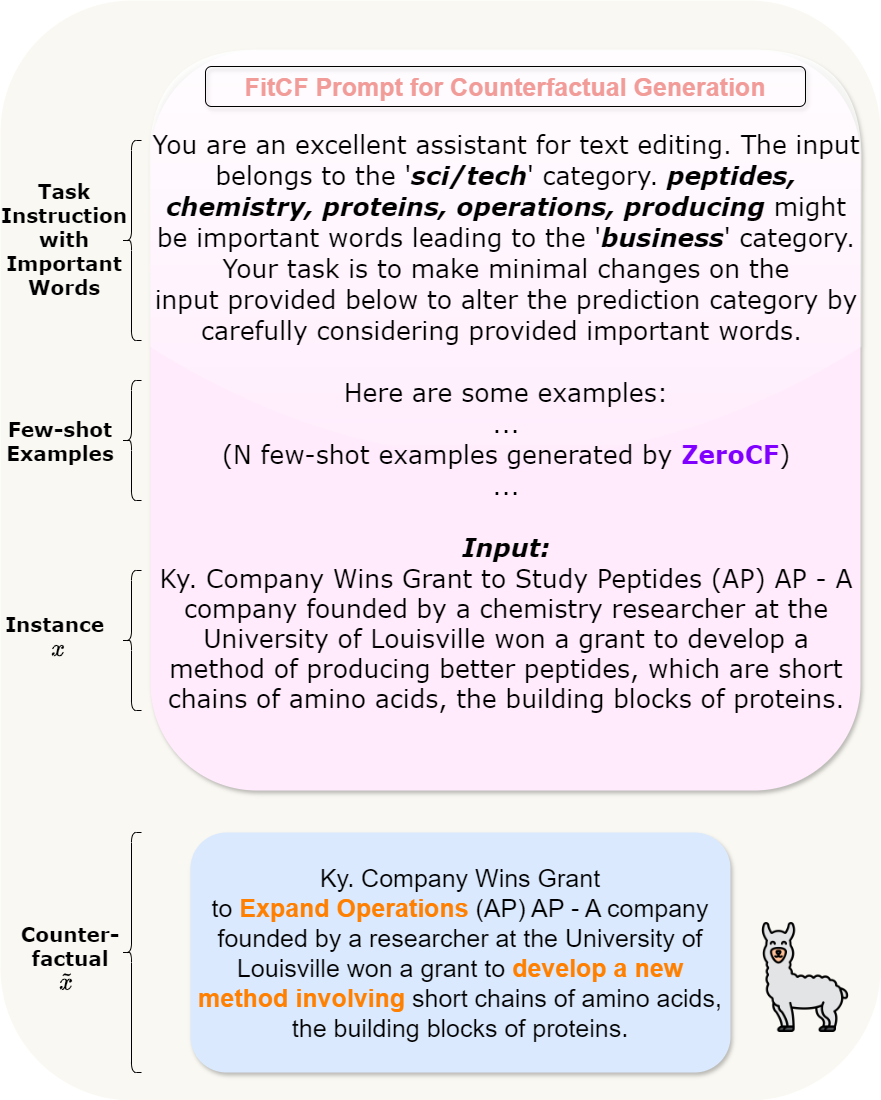}
\end{minipage}
}
\caption{Given an instance $x$ from the \data{AG News} dataset classified as ``\textbf{\textit{sci/tech}}'', our \sys{ZeroCF} approach generates few-shot examples, whose important words are determined by \textit{LIME} for a \lm{BERT} model. \sys{FitCF} then generates a counterfactual $\tilde{x}$ on this basis.
The edits to original instance $x$ are highlighted in orange, yielding $\tilde{x}$ which is classified as ``\textit{\textbf{business}}''.}
\label{fig:fitcf_example}
\end{figure}

\section{Introduction}
The advent of increasingly complex and opaque LLMs has triggered a critical need for explainability and interpretability of such models. Counterfactuals, which are minimally edited inputs that yield different predictions compared to reference inputs \cite{miller-2019-explanation, ross-etal-2021-explaining, madsen-2022-survey} are widely used in XAI and NLP. Applications include creating new data points for improving models in terms of performance \cite{kaushik-2020-learning, sachdeva-etal-2024-catfood} and robustness \cite{gardner-etal-2020-evaluating, ross-etal-2021-explaining}, and understanding the black-box nature of models \cite{wu-etal-2021-polyjuice, wang-etal-2024-llmcheckup, wang-etal-2024-coxql}. 
Crowd-sourcing counterfactuals can be costly, inefficient, and impractical \cite{chen-etal-2023-disco}, particularly in a specialized domain such as medicine. LLM-based counterfactual generation offers a more efficient and scalable alternative. Despite advancements in counterfactual generation techniques and the demonstrated versatility of LLMs across tasks \cite{wu-etal-2021-polyjuice, bhan-etal-2023-enhancing, li-etal-2024-prompting}, the efficacy of LLMs in producing high-quality counterfactuals in a zero-shot setting, as well as the effective construction of valid counterfactuals as demonstrations to enable few-shot prompting, remains an open question \cite{bhattacharjee-etal-2024-zero}. Additionally, the combination of widely used interpretability 
methods
, with the goal to exploit their combined benefits, has been insufficiently explored within XAI research \cite{treviso-etal-2023-crest, baeumel-etal-2023-investigating, bhan-etal-2023-enhancing}.

To this end, we first present \sys{ZeroCF}, a method to combine feature importance with counterfactual generation by leveraging important words identified through feature attribution scores for a fine-tuned \lm{BERT} \cite{devlin-etal-2019-bert} on the target dataset, evaluated on four representative feature importance methods (\S\ref{subsec:feature_importance}). The generation of counterfactuals with \sys{ZeroCF} is performed by prompting LLMs with extracted important words in a zero-shot setting without any auxiliary counterfactual data (\S\ref{subsec:zerocf_pipeline}). 
We then propose the \sys{FitCF} framework (Figure~\ref{fig:fitcf_example}), which uses \sys{ZeroCF}-generated counterfactuals following a label flip verification step as demonstrations for few-shot prompting without relying on human-crafted examples (\S\ref{subsec:fitcf_pipeline}).

Secondly, we evaluate \sys{ZeroCF} and \sys{FitCF} on two NLP tasks - news topic classification and sentiment analysis - using three baselines, \sys{Polyjuice} \cite{wu-etal-2021-polyjuice}, \sys{BAE} \cite{garg-ramakrishnan-2020-bae} and \sys{FIZLE} \cite{bhattacharjee-etal-2024-zero}. The automatic evaluation employs three automated metrics: Label flip rate, fluency, and edit distance. Both \sys{ZeroCF} and \sys{FitCF} significantly outperform \sys{Polyjuice} and \sys{BAE}, with \sys{ZeroCF} surpassing \sys{FIZLE} in most cases and \sys{FitCF} consistently exceeding three state-of-the-art baselines and \sys{ZeroCF}.

Thirdly, we perform ablation studies on three key components of \sys{FitCF}: (1) Important words; (2) the number of demonstrations; (3) label flip verification. The results reveal that all three components contribute positively to improving the quality of counterfactuals, as measured by label flip rate, fluency, and edit distance, with the number of demonstrations being the most influential. In addition, \sys{FitCF} exhibits greater robustness and achieves higher overall quality when combined with \textit{LIME} and \textit{SHAP} compared to its combination with \textit{Gradient} and \textit{Integrated Gradients}. 

Lastly, we conduct a correlation analysis between the quality of generated counterfactuals and the faithfulness of feature attribution scores as used in \sys{ZeroCF} and \sys{FitCF}. The analysis reveals that \textit{LIME} and \textit{SHAP} can produce more faithful feature attribution scores compared to \textit{Gradient} and \textit{Integrated Gradients}. Furthermore, we observe a strong correlation between the faithfulness of these feature attribution scores and the quality of counterfactuals generated by \sys{FitCF}.

\begin{figure*}[ht!]
\centering
\resizebox{\textwidth}{!}{
\begin{minipage}{\columnwidth}
\includegraphics[width=\columnwidth]{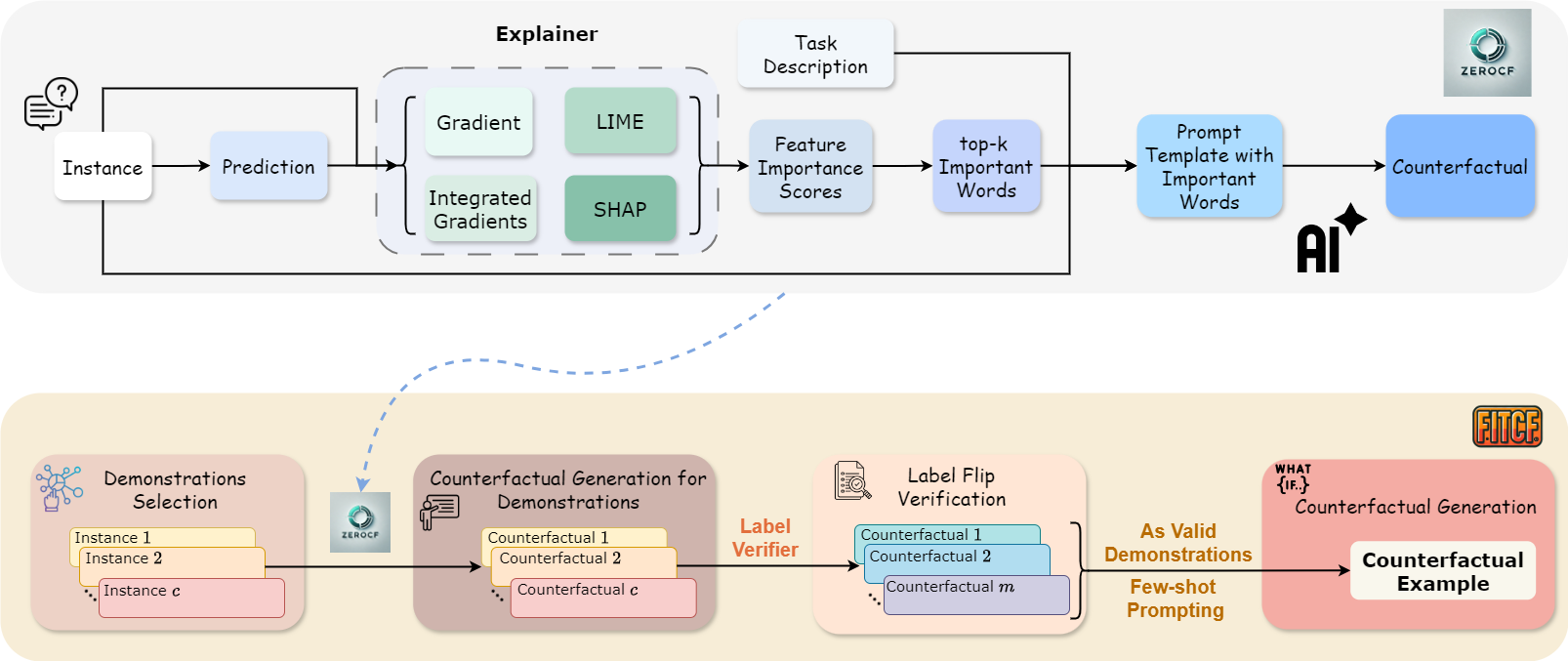}
\end{minipage}
}
\caption{The upper part of the figure illustrates how counterfactuals are generated by \sys{ZeroCF} using important words extracted by the explainer (\lm{BERT}) through various feature important methods (\textit{Gradient, Integrated Gradients, LIME, SHAP}). Lower part of the figure shows the pipeline of \sys{FitCF} involving demonstrations selection, automatic construction of counterfactual examples by \sys{ZeroCF}, label flip verification, and counterfactual generation.}
\label{fig:pipeline}
\end{figure*}


\section{Related Work}
\paragraph{Counterfactual Generation}
\sys{MICE} generates contrastive edits that change the prediction to a given contrast prediction \cite{ross-etal-2021-explaining}. \sys{Polyjuice} uses a fine-tuned \lm{GPT-2} \cite{radford-2019-language} to specify the type of edit needed to generate counterfactual examples \cite{wu-etal-2021-polyjuice}. \sys{DISCO} \cite{chen-etal-2023-disco} uses the \lm{GPT-3} fill-in-the-blank mode, which is not available in most open-source LLMs \cite{chen-etal-2023-disco}. \sys{Tigtec} \cite{bhan-et-al-2023-tictec} utilizes local feature importance to identify words that significantly influence a model's prediction and masks these important words using a fine-tuned masking model. \citet{bhattacharjee2024llmguidedcausalexplainabilityblackbox} identify the latent features in the input text and the input features associated with the latent features to generate counterfactual examples, which is criticized due to the additional level of complexity with no significant performance gain \cite{delaunay-2024-doesmakesenseexplain}. \sys{FIZLE} \cite{bhattacharjee-etal-2024-zero} shares the most similarity with \sys{FitCF} and uses LLMs as pseudo-oracles to generate counterfactuals with the assistance of LLM-generated important words in a zero-shot setting. 




\paragraph{Combination of Interpretability Methods} Recent works have explored the possibility to combine different XAI methods.
\citet{wang-2021-feature} propose a feature importance-aware attack, which disrupts important features that consistently influence the model's decisions. 
\citet{gressel-2023-featureimportanceguidedattack} identify perturbations in the feature space to produce evasion attacks. \citet{treviso-etal-2023-crest} present the framework, CREST, to generate counterfactual examples by combining rationalization with span-level masked language modeling. \citet{krishna-2023-amplify} employ various post-hoc explanations for rationalization, extending beyond counterfactuals, in contrast to CREST. 
\citet{bhan-etal-2023-enhancing} propose a method to determine impactful input tokens with respect to generated counterfactual examples. In contrast, \sys{FitCF} uses feature importance to guide counterfactual example generation.

\section{Methodology}
\label{sec:methodology}

\subsection{\sys{ZeroCF}}
\label{subsec:zerocf_pipeline}

\citet{bhattacharjee-etal-2024-zero} introduced \sys{FIZLE}, which generates counterfactuals in a zero-shot setting by prompting the LLM with important words identified by the LLM itself. However, these extracted words may be unfaithful or hallucinated \cite{li-2023-surveylargelanguagemodels}\footnote{Applying \lm{Llama3-8B} with \sys{FIZLE} on \data{AG News}, we find that for 64.5\% of the instances, a subset of generated important words is hallucinated, i.e., absent from the original input.}. To address this limitation, we propose \sys{ZeroCF} (Figure~\ref{fig:pipeline}; examples are provided in Table~\ref{tab:counterfactual_examples}), which relies on the most attributed words based on feature attribution scores determined by various explanation methods for the predictions of a \lm{BERT} model fine-tuned on the target dataset. 
Feature importance involves determining how significant an input feature is for a given output \cite{madsen-2022-survey}, which we find to enhance the counterfactual generation process (\S\ref{subsec:result_automatic_evaluation}). 

\paragraph{Prediction} Given an input $x$ from the dataset $\mathcal{D}$, we leverage a \lm{BERT} model $\mathcal{M_D}$ fine-tuned on $\mathcal{D}$\footnote{Detailed information, e.g., accuracy, about the deployed \lm{BERT} models is provided in Appendix~\ref{app:bert}.} to obtain the prediction $y_{pred}$ for the given input $x$:
\begin{equation}
    y_{pred} = \mathcal{M_D} (x)
\end{equation}

\paragraph{Feature Attribution Scores} Then we deploy an explainer $\mathcal{E}$ with access to the model $\mathcal{M_D}$, which employs various feature importance methods $f$ (\S\ref{subsec:feature_importance}) to acquire feature attributions scores $s$ based on the prediction $y_{pred}$ and the given input $x$:
\begin{equation}
    s = \mathcal{E}(x, y_{pred}, f, \mathcal{M_D})
\end{equation}

\paragraph{Counterfactual Generation} Finally, we identify the top-attributed words\footnote{The top attributed words are further post-processed by replacing the ``[CLS]'' and ``[SEP]'' special tokens if any, with the subsequent attributed words and by merging tokenized subwords if one of them is a top attributed word.} $w$ based on feature attribution scores $s$ and deploy an LLM $\mathcal{L}$ in a zero-shot setting to generate the counterfactual $\tilde{x}$ with the prompt $p$ (\S\ref{app:prompt_zerocf}), which consists of task instruction $i$, words $w$, the prediction $y_{pred}$, and the input $x$:
\begin{equation}
    \tilde{x} = \mathcal{L} (p)
\end{equation}

\subsection{\sys{FitCF}}
\label{subsec:fitcf_pipeline}

While \sys{ZeroCF} mitigates the issue of hallucinated important words extracted by the LLM, the counterfactuals generated by \sys{ZeroCF} may fail to flip the prediction, e.g., due to the limited capability of zero-shot prompting \cite{brown-2020-languagemodelsfewshotlearners}. To address it, we propose \sys{FitCF} (Figure~\ref{fig:fitcf_example}, Figure~\ref{fig:pipeline}), inspired by Auto-CoT \cite{zhang-2023-automatic}, which combines two interpretability methods, feature importance and counterfactual examples, leveraging their complementary advantages and automatically constructs demonstrations by \sys{ZeroCF} incorporating label-flip verification. Verified demonstrations subsequently enable few-shot prompting in \sys{FitCF}.

\paragraph{top-$k$ Examples Sampling} In order to diversify demonstration selection \cite{an-etal-2023-context,zhang-2023-automatic} and construct demonstrations automatically, we first convert all instances from the dataset $\mathcal{D}$ into sentence embeddings using \lm{SBERT}\footnote{\url{https://huggingface.co/sentence-transformers/all-mpnet-base-v2}}, and then apply $k$-means clustering on these sentence embeddings to form $k$ clusters\footnote{Clustering visualizations are given in Appendix~\ref{app:clustering_visualization}.}, where $k$ does not necessarily correspond to the exact number of predefined dataset labels. Afterwards, we select a total of $c$ instances which are closest to the centroid of each cluster\footnote{Selected examples and their corresponding counterfactuals for a given instance are provided in Appendix~\ref{app:demonstration_selection}.}. In such a way, we diversify the demonstrations, potentially mitigating any misleading effects caused by \sys{ZeroCF}, which may produce flawed counterfactuals. Finally, \sys{ZeroCF} is employed to generate counterfactuals for the $c$ selected instances using
simple heuristics.

\paragraph{Label Flip Verification} 

Subsequently, in order to validate the generated counterfactuals and to prevent incorrect counterfactuals from misleading the LLM \cite{turpin2023language}, we employ the same \lm{BERT} model $\mathcal{M_D}$ (\S\ref{subsec:zerocf_pipeline}) to make predictions on $c$ generated counterfactuals $\mathcal{C} = \{\tilde{x}_1, \tilde{x}_2, ..., \tilde{x}_c\}$ and the original input $\mathcal{X} = \{x_1, x_2, ..., x_c\}$ individually and assess whether the labels are inconsistent:
\begin{equation}
    \forall i \in \{1, 2, \cdots, c\}: \hat{y}_{x_i} = \mathcal{M_D} (x_i) 
\end{equation}
\begin{equation}
    \forall i \in \{1, 2, \cdots, c\}: \hat{y}_{\tilde{x}_i} = \mathcal{M_D} (\tilde{x}_i) 
\end{equation}

The generated counterfactuals $\tilde{x}_i$, where the predicted labels remain consistent $\hat{y}_{\tilde{x}_i} = \hat{y}_{x_i}$, are excluded from the demonstrations for further process to ensure the validity of the generated counterfactuals. In the end, we obtain $m$ counterfactuals, where $m \leq c$. To maintain a consistent number of demonstrations ($\ell$) for each input, if $m < \ell$, additional examples are iteratively selected based on their proximity to the cluster centroid, until the required number of demonstrations is achieved. Moreover, given we select $\ell$ demonstrations with $\ell >> k$, the influence of the number of clusters, $k$, on the final performance is diminished. 

\paragraph{Counterfactual Generation} For a given input $x$, $\ell$ input-counterfactual pairs generated by \sys{ZeroCF} are used as demonstrations, along with important words $w$ extracted based on the feature attribution scores $s$ generated by \lm{BERT} (\S\ref{subsec:zerocf_pipeline}), to prompt the LLM to generate the counterfactual for the input $x$ in a few-shot setting (Figure~\ref{fig:pipeline}, \S\ref{app:prompt_fitcf}).

\subsection{Considerations for Choice of Models}
In \sys{ZeroCF}, feature attributions are generated for a \lm{BERT} model's predictions, based on which important words are then extracted(\S\ref{subsec:zerocf_pipeline}). Moreover, in \sys{FitCF}, the same \lm{BERT} model serves as a label flip verifier (\S\ref{subsec:fitcf_pipeline}). \lm{BERT} is chosen as our design choice, because it is a representative encoder language model with a strong efficiency-performance balance. We emphasize that any model capable of performing classification tasks effectively can be used as a label flip verifier or for generating feature attribution scores. For encoder-only architectures like the \lm{BERT} model employed in our study, tools like \sys{Ferret} \cite{attanasio-etal-2023-ferret} can be used to derive feature attribution scores (\S\ref{subsec:feature_importance}). For encoder-decoder or decoder-only architectures, tools like \sys{Inseq} \cite{sarti-etal-2023-inseq} can generate such scores.

\section{Experimental Setup}

\subsection{Baselines}
We employ three approaches as baselines for \sys{ZeroCF} and \sys{FitCF}, including \sys{FIZLE} that achieves state-of-the-art performance in zero-shot counterfactual generation, which we aim to enhance.

\paragraph{BAE} \sys{BAE} is an adversarial attack method that employs \lm{BERT} to perturb input text by replacing masked words \cite{garg-ramakrishnan-2020-bae}.

\paragraph{Polyjuice} \sys{Polyjuice} allows users to control perturbation types and deploys a \lm{GPT-2}\footnote{Although \sys{Polyjuice} utilizes a relatively small model, \lm{GPT-2}, for generating counterfactuals, and \sys{BAE} uses \lm{BERT} to replace words based on embeddings, we consider both of them suitable baseline methods for \sys{FitCF}. This is because the deployed \lm{GPT-2} is \textbf{fine-tuned} on a counterfactual example dataset. Furthermore, \sys{FIZLE} relies on \textbf{zero-shot} prompting and achieves \textbf{state-of-the-art} performance.} to generate counterfactuals by framing the task as a conditional text
generation problem \cite{wu-etal-2021-polyjuice}.


\paragraph{FIZLE} \sys{FIZLE} employs an LLM to identify important words and prompts the LLM with these words in a zero-shot setting to generate counterfactuals \cite{bhattacharjee-etal-2024-zero}.

\subsection{Dataset}
\label{subsec:dataset}
Following \citet{nguyen-etal-2024-llms, bhattacharjee-etal-2024-zero}, we demonstrate the validity of \sys{ZeroCF} and \sys{FitCF} by applying them to two NLP tasks: News topic classification and sentiment analysis\footnote{Details on label distributions and example instances from the datasets used can be found in Appendix~\ref{app:dataset}.}.

\paragraph{AG News} \data{AG News} \cite{zhang-2015-agnews} contains news articles created by combining the titles and description fields of articles from four categories: \textit{World}, \textit{Sports}, \textit{Business}, and \textit{Sci/Tech}.

\paragraph{SST2} \data{SST2} \cite{socher-etal-2013-recursive} is part of the larger Stanford Sentiment Treebank and focuses specifically on binary sentiment classification of natural language movie reviews. Each sentence is labeled as either \textit{negative} or \textit{positive}.

\subsection{Models for Counterfactual generation}
We select three open source state-of-the-art instruction fine-tuned LLMs with increasing parameter sizes\footnote{More details about deployed models and inference time are provided in Appendix~\ref{app:experiment}.}:
\lm{Llama3-8B} \cite{llama3modelcard}, and \lm{Qwen2.5-\{32B,72B\}} \cite{qwen2.5}.

\subsection{Feature Importance}
\label{subsec:feature_importance}

\sys{Ferret} \cite{attanasio-etal-2023-ferret} is a framework that provides post-hoc explanations for LLMs and can evaluate these explanations based on faithfulness and plausibility. We use \sys{ferret} to generate feature attribution scores, selecting the following feature importance methods $f$: \textit{Gradient} \cite{simonyan-2014-inputgradient}, \textit{LIME} \cite{ribeiro-etal-2016-trust}, \textit{Integrated Gradients} \cite{sundararajan-2017-ig}, and \textit{SHAP} \cite{lundberg-etal-2017-shap}.

\section{Evaluation}

\subsection{Automatic Evaluation}
\label{subsec:automatic_evaluation}

The generated counterfactuals are evaluated using the following three automated metrics.

\paragraph{Soft Label Flip Rate} The Soft Label Flip Rate (SLFR) measures the frequency at which newly perturbed examples alter the original label to a different label \cite{ge-2021-counterfactualevaluationexplainableai, nguyen-etal-2024-llms, bhattacharjee2024llmguidedcausalexplainabilityblackbox}. For a dataset with $N$ instances, we calculate SLFR as follows:
\begin{equation*}
    SLFR = \frac{1}{N}\sum_{n=1}^{N} \mathds{1} (y_{k}^{'} \neq y_{k})
\end{equation*}
where $\mathds{1}$ is the indicator function, $y_{k}$ is the original label and $y_{k}^{'}$ is the predicted label after the perturbation. Note that we use \textbf{the same LLM} for both counterfactual generation and classification\footnote{The accuracy and error rate of the deployed LLMs, along with the prompt instruction used are provided in Appendix~\ref{app:calculation_lfr}.}. 

\paragraph{Perplexity} Perplexity (PPL) is defined as the exponential of the average negative log-likelihood of a sequence. PPL can measure the naturalness of the text distribution and how fluently the model can output the next word given the previous words \cite{fan-etal-2018-hierarchical}. Given a sequence $X = (x_0, x_1, \cdots, x_t)$, PPL of $X$ is calculated as:
\begin{equation*}
    PPL(X) = \exp\left\{\frac{1}{t}\sum_i^{t} \log{p_{\theta}(x_i|x_{<i})}\right\}
\end{equation*}
Following \citet{wang-2023-perplexityplmunreliableevaluating, nguyen-etal-2024-llms, bhattacharjee-etal-2024-zero}, we deploy \lm{GPT-2} to calculate PPL in our experiments due to its proven effectiveness in capturing such text distributions.

\paragraph{Textual Similarity (TS)} The counterfactual $\tilde{x}$ should be as similar as the original input $x$ \cite{madaan-2021-generate}, where lower distances indicate greater similarity. We use normalized word-level Levenshtein distances $d$ to capture all edits, which is widely used by the research community \cite{ross-etal-2021-explaining, treviso-etal-2023-crest}:
\begin{equation*}
    TS = \frac{1}{N} \sum_{i=1}^{N} \frac{d(x_i, \tilde{x}_i)}{|x_i|}
\end{equation*}

\subsection{Ablation Study}
As illustrated in Figure~\ref{fig:pipeline}, \sys{FitCF} comprises three core components: \textit{Important words}; \textit{demonstrations}; and \textit{label flip verification}. Accordingly, we conduct a comprehensive ablation study to evaluate the importance of each component individually. The experiments are conducted using \lm{Qwen2.5-72B}, as \lm{Qwen2.5-72B} particularly struggles to generate high-quality counterfactual examples compared to \lm{Llama3-8B} and \lm{Qwen2.5-32B} (Table~\ref{tab:automatic_evaluation}, Table~\ref{tab:different_number_of_demo}).

\subsubsection{Effect of Important Words}
To assess the contribution of important words identified by \lm{BERT} using different feature importance methods to counterfactual generation, we conduct the experiment using \sys{FitCF} omitting any pre-identified important words.

\begin{table*}[ht!]
    \centering
    \setlength{\extrarowheight}{3pt}
    \renewcommand*{\arraystretch}{0.7}
    
    \footnotesize
    \resizebox{0.95\textwidth}{!}{%
        \begin{tabular}{ccc|ccc|ccc}

        \toprule
         \multirow{2}{*}{\textbf{Approach}} & \multicolumn{2}{c|}{\textbf{Dataset}} & \multicolumn{3}{c|}{\textbf{\data{AG News} \scriptsize{($PPL=95.72$)}}} & \multicolumn{3}{c}{\textbf{\data{SST2} \scriptsize{($PPL=309.53$)}}}\\
         & \textbf{Model} & \textbf{Method} & SLFR $\uparrow$ & PPL $\downarrow$ & TS $\downarrow$ & SLFR $\uparrow$ & PPL $\downarrow$ & TS $\downarrow$\\

        \midrule

        \small{\sys{Polyjuice}} & \small{\lm{GPT2}}  &  - 
            & 18.60\% & 121.76 & 0.50
            & 29.00\% & 258.32 & 0.71 \\

        \cmidrule(lr){2-9}

         \sys{BAE} & \small{\lm{BERT}} &  - 
            & 19.50\% & 168.44 & 0.12
            & 47.00\% & 367.06 & 0.09 \\
            
        \cmidrule(lr){2-9}

        \multirow{3}{*}{\sys{FIZLE}} & \small{\lm{Llama3-8B}}   & -  & 93.50\% & 123.67 & 0.61
            & 95.50\% & 202.22 & 0.52 \\ 
        & \small{\lm{Qwen2.5-32B}} &  -   & 49.00\% & 53.07 & 1.14
            & 86.80\% & 167.51 & 0.66 \\ 
        & \small{\lm{Qwen2.5-72B}}   & -  & 21.50\% & 84.09 & 0.22 & 92.00\% & 257.91 & 0.43 \\ 

        \midrule

        \multirow{12}{*}{\small{\sys{ZeroCF} (Ours)}} & \multirow{4}{*}{\small{\lm{Llama3-8B}}} & Gradient & 93.50\% & 102.56 & 0.38 & 97.50\% & 239.15 & 0.46 \\ 

        & & IG
            & 95.50\% & 109.09 & 0.27 
            & \textbf{99.50\%} & 222.51 & \textbf{0.42}\\

        & &  LIME
            & 97.50\% & 107.72 & 0.39 
            & 97.00\% & 264.91 & 0.42\\ 
        & &  SHAP
            & \textbf{98.00\%} & \textbf{99.08} & \textbf{0.27} 
            & 94.00\% & \textbf{204.76} & 0.46\\ 

        \cmidrule(lr){2-9}

         & \multirow{4}{*}{\small{\lm{Qwen2.5-32B}}} & Gradient & \textbf{68.00\%} & 62.63 & 2.10 & 70.50\% & 205.06 & \textbf{0.48} \\ 

        & & IG
            & 51.00\% & \textbf{60.45} & \textbf{0.76}
            & 91.00\% & 222.57 & 0.64\\

        & &  LIME
            & 56.00\% & 63.75 & 0.84
            & 90.50\% & 576.59 & 0.62\\ 
        & &  SHAP
            & 55.50\% & 61.68 & 0.79
            & \textbf{93.00\%} & \textbf{191.00} & 0.60\\ 
            
        \cmidrule(lr){2-9} 
         
         & \multirow{4}{*}{\small{\lm{Qwen2.5-72B}}} & Gradient & 16.67\% & 74.19 & \textbf{0.21} 
            & 88.50\% & 263.47 & 0.34 \\ 

        & & IG
             & 24.50\% & 92.47 & 0.22
        & \textbf{92.00\%} & \textbf{281.10} & 0.46\\

        & &  LIME
            & 23.00\% & \textbf{72.73} & 0.71
            & 85.00\% & 289.20 & 0.30\\ 
        & &  SHAP
            & \textbf{25.00\%} & 73.92 & 0.74
            & 86.50\% & 319.60 & \textbf{0.22}\\ 

        \midrule


        \multirow{12}{*}{\small{\sys{FitCF} (Ours)}} & \multirow{4}{*}{\small{\lm{Llama3-8B}}} & Gradient & 94.50\% & 86.90 & 0.21  & 99.80\% & 159.57 & \textbf{0.47} \\ 

        & & IG
            & \textbf{96.00\%} & 87.67 & 0.23 
            & 100.00\% & 161.88 & 0.48\\

        & &  LIME
           & 95.50\% & \textbf{75.15} & \textbf{0.19} 
           & \textbf{100.00\%} & \textbf{151.22} & 0.48\\ 
        & &  SHAP
            & 94.00\% & 260.57 & 0.21
           & 100.00\% & 157.36 & 0.49\\ 

        \cmidrule(lr){2-9}
        
         & \multirow{4}{*}{\small{\lm{Qwen2.5-32B}}} & Gradient & 56.00\% & 62.97 & 0.73
            & 89.00\% & 214.25 & 0.51 \\ 

        & & IG
           & 57.50\% & \textbf{57.01} & \textbf{0.68} 
            & \textbf{90.50\%} & 221.64 & \textbf{0.49}\\

        & &  LIME
            & 56.00\% & 57.45 & 0.79
           & 89.50\% & 174.34 & 0.52\\ 
        & &  SHAP
            & \textbf{62.00\%} & 57.64 & 0.78
           & 89.50\% & \textbf{157.09} & 0.52\\ 
            
        \cmidrule(lr){2-9} 
         
         & \multirow{4}{*}{\small{\lm{Qwen2.5-72B}}} & Gradient & \textbf{77.00\%} & 62.13 & 0.99
            & 96.00\% & 595.71 & \textbf{0.38} \\ 

        & & IG
             & 42.00\% & 63.54 & \textbf{0.33}
            & 95.00\% & \textbf{207.55} & 0.39\\

        & &  LIME
            & 45.00\% & \textbf{61.54} & 0.35
           & \textbf{96.50\%} & 240.94 & 0.41\\ 
        & &  SHAP
            & 38.96\% & 67.28 & 0.34
           & 96.50\% & 590.94 & 0.39\\

        \bottomrule
        \end{tabular}
    }
    \caption{Automatic evaluation results of counterfactuals generated by \sys{Polyjuice}, \sys{BAE}, \sys{FIZLE}, \sys{ZeroCF}, and \sys{FitCF} with \lm{Llama3-8B}, \lm{Qwen2.5-32B}, and \lm{Qwen2.5-72B} using Soft Label Flip Rate (SLFR), Perplexity (PPL), and Textual Similarity (TS) on \data{AG News} and \data{SST2}. Bold faced values indicate for each approach, which feature importance method is the best performing according to the respective metric.
    }
    \label{tab:automatic_evaluation}
\end{table*}

\subsubsection{Effect of Number of Demonstrations}
\label{subsubsec:effect_of_num_demo}
In \sys{FitCF}, as $k$ clusters are obtained through clustering, and due to the difficulty and complexity of counterfactual example generation, we set the number of demonstrations to twice the number of clusters for each dataset ($\ell=2k$; \S\ref{subsec:fitcf_pipeline}), which results in 10 demonstrations for \data{AG News} and 8 for \data{SST2}, respectively (Figure~\ref{fig:clustering_visualization}). To examine the effect of the number of demonstrations and assess the necessity of doubling the number of demonstrations to $2k$, we further evaluate the quality of counterfactual examples generated by \sys{FitCF}, with the number of demonstrations set to the number of clusters ($\ell=k$). \looseness=-1

\subsubsection{Effect of Label Flip Verification}
To ensure the validity of the selected demonstrations and prevent incorrect examples from misleading the LLM \cite{rubin-etal-2022-learning,turpin2023language}, \sys{FitCF} incorporates a label flip verifier (\S\ref{subsec:fitcf_pipeline}). This verifier is implemented using a fine-tuned \lm{BERT} model (Table~\ref{tab:bert_model}) trained on the target dataset. To assess the impact of label flip verification, we conduct an ablation study by excluding label flip verification for comparative analysis.

\subsection{Correlation Analysis}
As we deploy various feature importance methods to generate counterfactuals synergistically (Figure~\ref{fig:pipeline}), which can then be applied as demonstrations in \sys{FitCF}, we investigate the correlation between the quality of the feature attribution scores and the quality of generated counterfactuals. The feature attribution scores are evaluated based on faithfulness using \sys{ferret} \cite{attanasio-etal-2023-ferret}. For faithfulness evaluation, we employ three metrics:  \textit{comprehensiveness}, \textit{sufficiency} \cite{deyoung-etal-2020-eraser} and \textit{Kendall's $\tau$ correlation with Leave-One-Out token removal} \cite{jain-wallace-2019-attention}.

\begin{table}[t!]
    \centering
    \setlength{\extrarowheight}{3pt}
    \renewcommand*{\arraystretch}{0.8}
    \footnotesize
    \resizebox{\columnwidth}{!}{%
        \begin{tabular}{c|c|ccc}

        \toprule
          \textbf{Dataset} & \textbf{Method} & \textbf{SLFR} & \textbf{PPL} & \textbf{TS}\\

        \midrule

        \multirow{4}{*}{\centering \rotatebox[origin=c]{90}{\small{\data{AG News}}}}  & Gradient
            & 41.50\% \color{ForestGreen}($\downarrow$\color{ForestGreen}{35.50}\%) & 67.85 \color{ForestGreen}($\downarrow$\color{ForestGreen}{5.72}) & 0.36 \color{red}($\uparrow$\color{red}0.63)\\ 

        & IG 
            & 37.50\% \color{ForestGreen}($\downarrow$\color{ForestGreen}{4.50}\%) &  67.85 \color{ForestGreen}($\downarrow$\color{ForestGreen}{4.31}) & 0.37 \color{red}($\uparrow$\color{red}0.62)\\ 

         & LIME 
           & 40.68\% \color{ForestGreen}($\downarrow$\color{ForestGreen}{4.32}\%) &  66.08 \color{ForestGreen}($\downarrow$\color{ForestGreen}{2.54}) & 0.35 \color{red}($\uparrow$\color{red}{0.02})\\ 
           
        & SHAP 
           & 37.00\% \color{ForestGreen}($\downarrow$\color{ForestGreen}{1.96}\%)  & 84.14 \color{ForestGreen}($\downarrow$\color{ForestGreen}16.86) & 0.51 \color{ForestGreen}($\downarrow$\color{ForestGreen} 0.17)\\ 

           \midrule

        \multirow{4}{*}{\centering \rotatebox[origin=c]{90}{\small{\data{SST2}}}}  & Gradient
            & 93.50\% \color{ForestGreen}($\downarrow$\color{ForestGreen}{2.50}\%) & 214.27 \color{red}($\uparrow$\color{red}{381.44}) & 0.42 \color{ForestGreen}($\downarrow$\color{ForestGreen} 0.04)\\ 

        & IG 
            & 95.00\% (- 0.00\%) &  214.27 \color{ForestGreen}($\downarrow$\color{ForestGreen}{6.72}) &  0.42 \color{ForestGreen}($\downarrow$\color{ForestGreen}{0.02})\\ 

         & LIME 
           & 95.50\% \color{ForestGreen}($\downarrow$\color{ForestGreen}{1.00}\%) & 278.78 \color{ForestGreen}($\downarrow$\color{ForestGreen}{37.84}) & 0.41 (-0.00)\\ 
           
        & SHAP 
           & 96.00\% \color{ForestGreen}($\downarrow$\color{ForestGreen}{0.50}\%)  & 290.57 \color{red}($\uparrow$\color{red}-300.37) & 0.43 \color{ForestGreen}($\downarrow$\color{ForestGreen}{0.04})\\

        \bottomrule
        \end{tabular}
    }
    \caption{Automatic evaluation results of counterfactuals generated by \sys{FitCF} using \lm{Qwen2.5-72B}, with demonstrations generated by \sys{ZeroCF} \underline{without} specifying \textit{important words}.
    }
    \label{tab:without_important_words}
\end{table}

\section{Results}
\subsection{Automatic Evaluation}
\label{subsec:result_automatic_evaluation}

Table~\ref{tab:automatic_evaluation} demonstrates that our proposed approaches, \sys{ZeroCF} and \sys{FitCF}, consistently outperform \sys{Polyjuice} and \sys{BAE} easily, which exhibit relatively low SLFR. Notably, \sys{BAE} achieves the lowest edit distance compared to \sys{Polyjuice}, \sys{FIZLE}, \sys{ZeroCF}, and \sys{FitCF}, as it only replaces masked words based on textual embeddings. For \data{AG News} dataset using \lm{Qwen2.5-32B}, the edit distance is comparatively higher than that of \sys{Polyjuice}, and the other baseline, \sys{FIZLE}, also shows a larger edit distance compared to \sys{Polyjuice}. For \data{SST2} dataset, \lm{Qwen2.5-72B} tends to generate counterfactuals that are less natural and fluent when leveraging \sys{ZeroCF} and \sys{FitCF}. Interestingly, \lm{Llama3-8B}, the smallest model among all evaluated LLMs, achieves the best overall performance. In contrast, \lm{Qwen2.5-72B} generally underperforms compared to both \lm{Llama3-8B} and \lm{Qwen2.5-32B}, as \lm{Qwen2.5-72B} has a stronger capability to discern the underlying context, making it less prone to flipping labels (Appendix~\ref{app:demonstration_selection}).

Additionally, we observe that \sys{ZeroCF} does not outperform \sys{FIZLE} in some cases, e.g., with \lm{Qwen2.5-72B} on \data{SST2} dataset. However, in most cases, \sys{ZeroCF} offers noticeable advantages in enhancing the quality of counterfactuals compared to \sys{FIZLE}. Furthermore, we find that \textit{Integrated Gradients} and \textit{SHAP} contribute more positively to the quality of counterfactuals, on average\footnote{We do not consider the number of times a feature importance method achieves the maximum value in tables, but rather the average ranking of a method across all datasets.}, compared to other feature importance methods. 

Importantly, \sys{FitCF} emerges as the most effective method for generating high-quality counterfactuals, consistently outperforming three baselines and \sys{ZeroCF} across all evaluated settings, underscoring its robustness and effectiveness. This demonstrates the advantage of combining feature importance with the counterfactual generation process. Under the \sys{FitCF} framework, \textit{Integrated Gradients} and \textit{LIME} illustrate superior performance in generating counterfactuals compared to the other two approaches.



\subsection{Ablation Study}
The results of the ablation studies are presented in Table~\ref{tab:without_important_words}, \ref{tab:different_number_of_demo}, \ref{tab:label_flip_verification}, where for PPL and TS, an upward (\textit{downward}) arrow signifies that a decrease (\textit{increase}) in the value corresponds to an improvement (\textit{deterioration}) in both metrics.

\subsubsection{Effect of Important Words}
\label{subsubsec:eval_demonstrations}

Table~\ref{tab:without_important_words} shows that for \data{AG News}, SLFR decreases across all methods, with the most significant decline observed when using \textit{Gradient}. Concurrently, PPL improves and edit distances generally increases, suggesting that the generated counterfactuals diverge more from the original text, except when using \textit{SHAP}. In contrast, for \data{SST2}, SLFR remains consistently high, with slight decreases. PPL exhibited mixed results, with both notable increases and decreases depending on the method, reflecting variability in fluency. Meanwhile, edit distance either decreases or remains unchanged. Overall, \sys{FitCF} with \textit{SHAP} demonstrates the highest robustness when important words are not specified, whereas \textit{Gradient} is particularly sensitive to the inclusion of important words.

\begin{table}[t!]
    \centering
    \setlength{\extrarowheight}{3pt}
    \renewcommand*{\arraystretch}{0.8}
    \footnotesize
    \resizebox{\columnwidth}{!}{%
        \begin{tabular}{c|c|ccc}

        \toprule
          \textbf{Dataset} & \textbf{Method} & \textbf{SLFR} & \textbf{PPL} & \textbf{TS}\\

        \midrule

        \multirow{4}{*}{\centering \rotatebox[origin=c]{90}{\small{\data{AG News}}}}  & Gradient
            & 13.50\% \color{ForestGreen}($\downarrow$\color{ForestGreen}{63.50}\%) & 66.74 \color{ForestGreen}($\downarrow$\color{ForestGreen}{4.61}) & 0.27 \color{red}($\uparrow$\color{red}0.72)\\ 

        & IG 
            & 15.50\% \color{ForestGreen}($\downarrow$\color{ForestGreen}{22.00} \%) & 64.28 \color{ForestGreen}($\downarrow$\color{ForestGreen}{0.74}) & 0.27 \color{red}($\uparrow$\color{red}0.06)\\ 

         & LIME 
           & 18.00\% \color{ForestGreen}($\downarrow$\color{ForestGreen}{27.00}\%) & 68.28 \color{ForestGreen}($\downarrow$\color{ForestGreen}{6.74}) & 0.27 \color{ForestGreen}($\downarrow$\color{ForestGreen}{0.08})\\ 
           
        & SHAP 
           & 14.00\% \color{ForestGreen}($\downarrow$\color{ForestGreen}{24.96}\%)  & 64.06 \color{red}($\uparrow$\color{red}3.22) & 0.28 \color{red}($\uparrow$\color{red}0.06)\\ 

           \midrule

        \multirow{4}{*}{\centering \rotatebox[origin=c]{90}{\small{\data{SST2}}}}  & Gradient
            & 89.00\% \color{ForestGreen}($\downarrow$\color{ForestGreen}{7.00}\%) &  235.08 \color{red}($\uparrow$\color{red}{360.63}) & 0.36 \color{red}($\uparrow$\color{red}0.02)\\ 

        & IG 
            & 93.50\% \color{ForestGreen}($\downarrow$\color{ForestGreen}{1.50} \%) & 266.09 \color{ForestGreen}($\downarrow$\color{ForestGreen}{58.54}) & 0.39 (-0.00)\\ 

         & LIME 
           & 91.50\% \color{ForestGreen}($\downarrow$\color{ForestGreen}{5.00}\%) & 250.70 \color{ForestGreen}($\downarrow$\color{ForestGreen}{9.76}) & 0.39 \color{red}($\uparrow$\color{red}{0.02})\\ 
           
        & SHAP 
           & 92.00\% \color{ForestGreen}($\downarrow$\color{ForestGreen}{4.50}\%)  & 583.42 \color{red}($\uparrow$\color{red}7.52) & 0.38 \color{red}($\uparrow$\color{red}0.01)\\

        \bottomrule
        \end{tabular}
    }
    \caption{Automatic evaluation results of counterfactuals generated by \sys{FitCF} with \lm{Qwen2.5-72B} using $k$ demonstrations.
    }
    \label{tab:different_number_of_demo}
\end{table}

\subsubsection{Effect of Number of Demonstrations}

As shown in Table~\ref{tab:different_number_of_demo}, we find that the number of demonstrations plays an critical role in the performance of \sys{FitCF}. For \data{AG News}, SLFR declines precipitously when the number of clusters ($k$) is used as the number of demonstrations (\S\ref{subsubsec:effect_of_num_demo}), while the edit distance shows a slight improvement. In comparison, for \data{SST2}, the degree of SLFR diminishment is less conspicuous. 

Furthermore, Table~\ref{tab:different_number_of_demo} reveals that in general, \sys{FitCF} with \textit{Integrated Gradients} and \textit{SHAP} exhibits greater robustness compared to \textit{Gradient} and \textit{LIME}. In particular, \sys{FitCF} with \textit{Gradient} demonstrates the highest sensitivity, with a strong decline in quality as the number of demonstrations decreases.

\subsubsection{Effect of Label Flip Verification}
\begin{table}[t!]
    \centering
    \setlength{\extrarowheight}{3pt}
    \renewcommand*{\arraystretch}{0.8}
    \footnotesize
    \resizebox{\columnwidth}{!}{%
        \begin{tabular}{c|c|ccc}

        \toprule
          \textbf{Dataset} & \textbf{Method} & \textbf{SLFR} & \textbf{PPL} & \textbf{TS}\\

        \midrule

        \multirow{4}{*}{\centering \rotatebox[origin=c]{90}{\small{\data{AG News}}}}  & Gradient
            & 34.00\% \color{ForestGreen}($\downarrow$\color{ForestGreen}{43.00}\%) &  63.27 \color{ForestGreen}($\downarrow$\color{ForestGreen}{1.14}) & 0.33 \color{red}($\uparrow$\color{red} 0.66)\\ 

        & IG 
            & 40.50\% \color{ForestGreen}($\downarrow$\color{ForestGreen}{1.50}\%) &  64.65 \color{ForestGreen}($\downarrow$\color{ForestGreen}{1.11}) & 0.35 \color{ForestGreen}($\downarrow$\color{ForestGreen} 0.02)\\ 

         & LIME 
           & 42.50\% \color{ForestGreen}($\downarrow$\color{ForestGreen}{2.50}\%) & 65.23 \color{ForestGreen}($\downarrow$\color{ForestGreen}3.69) & 0.35 (- 0.00)\\ 
           
        & SHAP 
           & 34.00\% \color{ForestGreen}($\downarrow$\color{ForestGreen}{4.96}\%)  & 65.30 \color{red}($\uparrow$\color{red}1.98) & 0.34 (- 0.00)\\ 

           \midrule

        \multirow{4}{*}{\centering \rotatebox[origin=c]{90}{\small{\data{SST2}}}}  & Gradient
            & 94.50\% \color{ForestGreen}($\downarrow$\color{ForestGreen}{1.50}\%) & 222.52 \color{red}($\uparrow$\color{red}{373.19}) &  0.36 \color{red}($\uparrow$\color{red}0.02)\\ 

        & IG 
            & 94.50\% \color{ForestGreen}($\downarrow$\color{ForestGreen}2.00\%) &  240.11 \color{ForestGreen}($\downarrow$\color{ForestGreen}{32.56}) & 0.39 (- 0.00)\\ 

         & LIME 
           & 96.00\% \color{ForestGreen}($\downarrow$\color{ForestGreen}{0.50}\%) & 245.79 \color{ForestGreen}($\downarrow$\color{ForestGreen}{4.85}) & 0.40 \color{red}($\uparrow$\color{red}{0.01})\\ 
           
        & SHAP 
           & 94.50\% \color{ForestGreen}($\downarrow$\color{ForestGreen}{2.00}\%)  & 281.65 \color{red}($\uparrow$\color{red}309.29) & 0.38 \color{red}($\uparrow$\color{red}{0.01})\\

        \bottomrule
        \end{tabular}
    }
    \caption{Automatic evaluation results of counterfactuals generated by \sys{FitCF} using \lm{Qwen2.5-72B}, \underline{without} \textit{label flip verification}.
    }
    \label{tab:label_flip_verification}
\end{table}
Table~\ref{tab:label_flip_verification} divulges trends similar to those observed in Table~\ref{tab:without_important_words} (\S\ref{subsubsec:eval_demonstrations}). Omitting label flip verification leads to decreases in SLFR across both datasets, highlighting the importance of this step. However, skipping label flip verification occasionally results in lower PPL for certain methods, suggesting improved fluency in some cases.

Meanwhile, the decrease in SLFR is more pronounced for \data{AG News}, particularly with the \textit{Gradient} method, which shows the largest SLFR drop alongside increases in PPL. Conversely, \textit{Integrated Gradients} and \textit{LIME} present minimal impact on SLFR, indicating a relative reliance on label flip verification to maintain consistent performance. 

\subsection{Discussion}
\label{subsub:discussion}
Important words identified through feature attribution scores for \lm{BERT} are more effective and less prone to hallucination for counterfactual generation compared to those self-generated by LLMs. Through ablation studies on the three core components of \sys{FitCF}, we conclude that the number of demonstrations generated by \sys{ZeroCF} has the most significant impact on the performance of \sys{FitCF}. While specifying important words and applying label flip verification also contribute to \sys{FitCF}'s effectiveness, their influence is less marked compared to the number of demonstrations. While SLFR decreases across three tables, the edit distance gets improved overall, except for \data{SST}, where no important words are specified. This indicates that without a certain component, the counterfactuals generated by \sys{FitCF} are generally less edited, resulting in less successful label flips. Moreover, \sys{FitCF} with \textit{Gradient} proves to be the least robust, showing substantial drops in SLFR, when any of the three components is removed. In contrast, \sys{FitCF} with \textit{LIME} and \textit{SHAP} demonstrate greater robustness and consistently produce high-quality counterfactuals. \looseness=-1


    \begin{table}[t!]
        \centering
        \setlength{\extrarowheight}{3pt}
        \renewcommand*{\arraystretch}{0.8}
        \footnotesize
        \resizebox{\columnwidth}{!}{%
            \begin{tabular}{p{0.3cm}|c|ccc|ccc}
    
            \toprule
              \multirow{2}{*}{\centering \tiny{\rotatebox[origin=c]{90}{\textbf{Model}}}} & \textbf{Dataset} & \multicolumn{3}{c|}{\textbf{\data{AG News}}} & \multicolumn{3}{c}{\textbf{\data{SST2}}}\\
             & Method & comp. & suff. & $\tau$ (loo) & comp. & suff. & $\tau$ (loo)\\
    
            \midrule
    
            \multirow{4}{*}{\centering \rotatebox[origin=c]{90}{\small{\lm{Llama3}}}} & Gradient
                & 0.20 & 0.13 & 0.06 & 0.21 & 0.25 & -0.03 \\ 
    
             & IG 
                & 0.38 & 0.03 & 0.07 & -0.52 & 0.05 & 0.22 \\  
    
             & LIME 
               & 0.61 & -0.02 & 0.16 & 0.68 & 0.02 & 0.29 \\ 
               
            & SHAP 
              & 0.62 & -0.02 & 0.16 & 0.60 & 0.03 & 0.25 \\
              
    
            \midrule
    
            \multirow{4}{*}{\centering \rotatebox[origin=c]{90}{\small{\lm{Qwen-32B}}}} &  Gradient
               & 0.12 & 0.12 & 0.07 & 0.20 & 0.23 & -0.03 \\ 
    
            & IG 
                & 0.32 & 0.03 & 0.05 & 0.50 & 0.04 & 0.21 \\  
    
             & LIME 
               & 0.53 & -0.01 & 0.12 & 0.67 & 0.01 & 0.29 \\ 
            & SHAP 
               & 0.53 & -0.01 & 0.08 & 0.59 & 0.02 & 0.25 \\  
    
            \midrule
    
            \multirow{4}{*}{\centering \rotatebox[origin=c]{90}{\small{\lm{Qwen-72B}}}} &  Gradient
               & 0.12 & 0.12 & 0.07 & 0.20 & 0.23 & -0.03\\ 
    
            & IG 
                & 0.32 & 0.03 & 0.05 & 0.50 & 0.04 & 0.21 \\ 
    
             & LIME 
               & 0.53 & -0.01 & 0.12 & 0.67 & 0.01 & 0.29 \\ 
               
           & SHAP 
               & 0.53 & -0.01 & 0.07 & 0.59 & 0.02 & 0.25 \\

            \bottomrule
            \end{tabular}
        }
        \caption{Faithfulness evaluation results based on \textit{Comprehensiveness} (comp.), \textit{Sufficiency} (suff.) and \textit{Kendall's $\tau$ correlation with Leave-One-Out token removal} ($\tau$ (loo)) for counterfactuals generated by \sys{FitCF} using \lm{Llama3-8B}, \lm{Qwen2.5-32B}, and \lm{Qwen2.5-72B} on \data{AG News} and \data{SST2} datasets.
        }
        \label{tab:correlation_analysis}
    \end{table}

\begin{figure*}[t!]
\centering
\resizebox{0.7\textwidth}{!}{
\begin{minipage}{\columnwidth}
\includegraphics[width=\columnwidth]{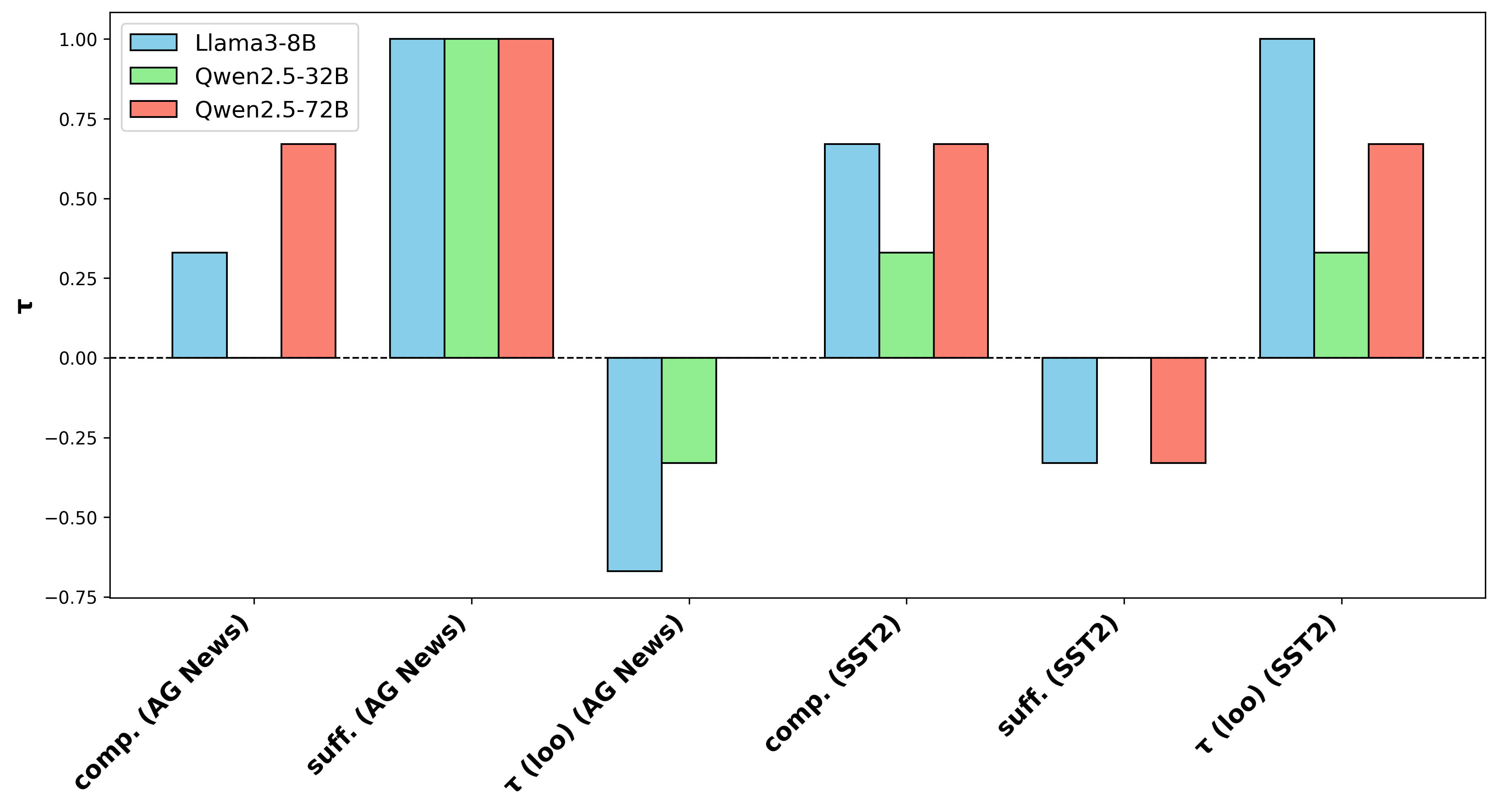}
\end{minipage}
}
\caption{A Kendall’s tau ($\tau$) that quantifies the degree of correspondence between the ranking of generated counterfactuals' \textit{quality} and the ranking of \textit{feature attribution evaluation results} is reported.}
\label{fig:tau}
\end{figure*}

\subsection{Correlation Analysis}
\label{subsec:eval_correlation_analysis}

From Table~\ref{tab:correlation_analysis}, we discover that \textit{LIME} and \textit{SHAP} consistently outperform \textit{Gradient} and \textit{Integrated Gradients} in terms of comprehensiveness and $\tau$ (loo) across all models and datasets, which aligns with our findings in \S\ref{subsub:discussion}. In addition, the comprehensiveness and sufficiency scores exhibit less variation across three models for \data{AG News}, though they are generally lower than those for \data{SST2}. In contrast, $\tau$ (loo) scores for \data{SST2} are slightly higher compared to \data{AG News}. Furthermore, for \data{AG News}, a strong correlation ($\tau=1$) is observed in Figure~\ref{fig:tau} between the quality of generated counterfactuals and sufficiency, while for \data{SST2}, 
both comprehensive and $\tau$ (loo) demonstrate notable correlations with counterfactual quality. We conclude that the faithfulness of feature attribution scores is generally strongly correlated with the quality of counterfactuals generated with the auxiliary assistance of extracted important words using \sys{FitCF}.

\section{Conclusion}
We first introduced \sys{ZeroCF}, an approach that leverages important words derived from feature attribution methods for counterfactual example generation in a zero-shot setting. Building on this, we proposed \sys{FitCF}, a framework that automatically constructs high-quality demonstrations using \sys{ZeroCF}, eliminating the need for human-annotated ground truth for counterfactual generation. \sys{FitCF} validates counterfactuals via label flip verification for their suitability as demonstrations in a few-shot setting. Empirically, \sys{FitCF} outperforms three state-of-the-art baselines \sys{Polyjuice}, \sys{BAE} and \sys{FIZLE}, and our own \sys{ZeroCF}. Through ablation studies, we identified the three core components of \sys{FitCF} - number of demonstrations, important words, and label flip verification - as critical to enhancing counterfactual quality. Moreover, we evaluated the faithfulness of feature attribution scores and found that \textit{LIME} and \textit{Integrated Gradients} are the most effective feature importance methods for \sys{FitCF}, consistently producing the most faithful feature attribution scores. Finally, our analysis revealed a strong correlation between the faithfulness of feature attribution scores and the quality of the generated counterfactuals.

Future work includes investigating the correlation between additional dimensions of feature attribution scores, such as \textit{plausibility}, \textit{coherence} and \textit{insightfulness}, and the quality of counterfactuals through user studies \cite{domnich2024unifyingevaluationcounterfactualexplanations}. We also plan to explore the potential of language models with architectures beyond encoder-only models as a foundation for feature attributions to be used in \sys{ZeroCF} and \sys{FitCF}. Furthermore, we would like to explore counterfactual example generation beyond typical text classification tasks to include, for instance, (long-form) question answering and open-ended mathematical problem, which are particularly challenging even for LLMs \cite{dehghanighobadi2025llmsexplaincounterfactually}.

\section*{Limitations}
We conducted experiments exclusively using datasets in English. In other languages, the current approach may not offer the same advantages.


In \sys{ZeroCF} and \sys{FitCF}, feature attribution scores are determined by an explanation method for the predictions of a \lm{BERT} model fine-tuned on the target dataset and the same \lm{BERT} model is used to verify label flips. The potential contribution of other language models to performing both tasks in \sys{ZeroCF} and \sys{FitCF}, however, remains unexplored. 

For feature attribution score evaluation, we focus only on faithfulness, although \sys{ferret} also supports the evaluation of plausibility. However, plausibility evaluation requires human annotations, so we have left it out and consider it as future work.


\section*{Acknowledgment} 
We are indebted to the anonymous reviewers of ACL 2025 for their helpful and rigorous feedback.
This work has been supported by the German Federal Ministry of Research, Technology and Space as part of the projects VERANDA (16KIS2047) and BIFOLD 24B.

\bibliography{custom}
\appendix

\section{Prompt Instruction}

\subsection{Prompt for \sys{ZeroCF}}
\label{app:prompt_zerocf}

\begin{verbatim}
You are an excellent assistant for text 
editing. You are given an input from the 
{dataset} dataset, classified into one of 
{len(labels)} categories: 
{', '.join(labels)}. The input belongs to 
the '{prediction}' category.
{important_words} might be important words 
leading to the '{prediction}' category. 

Your task is to make minimal changes on the 
below provided input to alter the 
prediction category by carefully 
considering provided important words. 
Please output only the edited input.

Input: {input_text}
\end{verbatim}

\subsection{Prompt for \sys{FitCF}}
\label{app:prompt_fitcf}

\begin{verbatim}
You are an excellent assistant for text 
editing. You are given an input from the 
{dataset} dataset, classified into one of 
{len(labels)} categories: 
{', '.join(labels)}. The input belongs to 
the '{prediction}' category.
{important_words} might be important words 
leading to the '{prediction}' category.

Your task is to make minimal changes on the 
input provided below to alter the 
prediction category to '{counterpart}' by 
carefully considering provided important 
words and examples. Please output the 
edited input only!

Below are some examples consisting of 
original and edited input.

[original input] {original_input_1}
[edit input] {edit_input_1}
...
[original input] {input_text}
[edit input] 
\end{verbatim}

\section{Detailed Information of Deployed \lm{BERT}}
\label{app:bert}
\begin{table*}[t!]
    \centering
    \resizebox{\textwidth}{!}{%
        \begin{tabular}{rccc}

        \toprule
        \textbf{Dataset} & \textbf{Model} & \textbf{Accuracy} & \textbf{Link}\\

        \midrule
        \data{AG News} & \lm{textattack/bert-base-uncased-ag-news} & 93.03\% & \url{https://huggingface.co/textattack/bert-base-uncased-ag-news}\\

        
        \data{SST2} & \lm{gchhablani/bert-base-cased-finetuned-sst2} & 92.32\% & \url{https://huggingface.co/gchhablani/bert-base-cased-finetuned-sst2}\\
        \bottomrule
        \end{tabular}
        }
    \caption{
    \lm{BERT} models used for \data{AG News} and \data{SST2} datasets, with accuracy validated on their respective testsets.
    }
    \label{tab:bert_model}
\end{table*}

Table~\ref{tab:bert_model} displays \lm{BERT} models used for \data{AG News} and \data{SST2} datasets with their validation accuracies. As both \lm{BERT} models demonstrate strong performance in accuracy, we can use them as classifiers (\S\ref{subsec:zerocf_pipeline}) and label flip verifiers (\S\ref{subsec:fitcf_pipeline}).

\section{Visualization of Clustering}
\label{app:clustering_visualization}
\begin{figure}
  \centering

  \begin{subfigure}{\columnwidth}
    \centering
    \centering
\includegraphics[width=\linewidth]{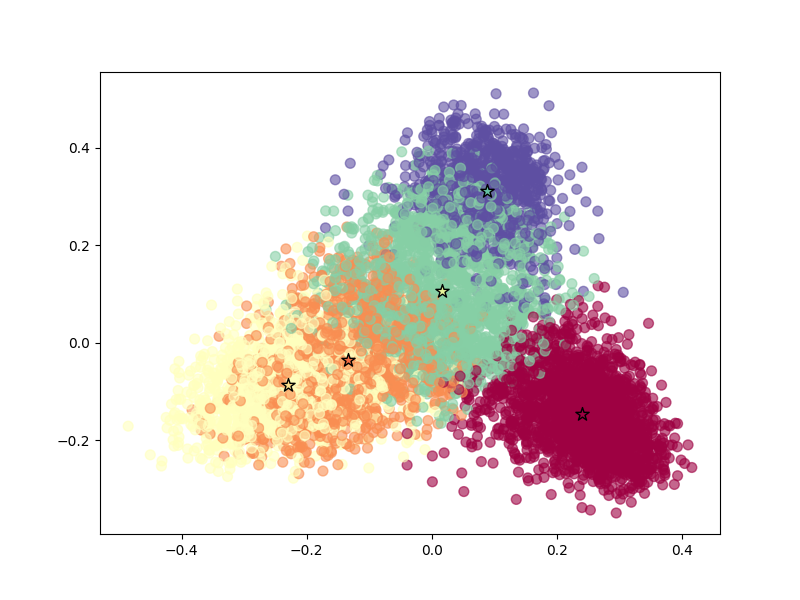}
\caption{\data{AG News}}
\label{fig:ag_news_visualization}
  \end{subfigure}
  \hfill
  \begin{subfigure}{\columnwidth}
\centering
\includegraphics[width=\linewidth]{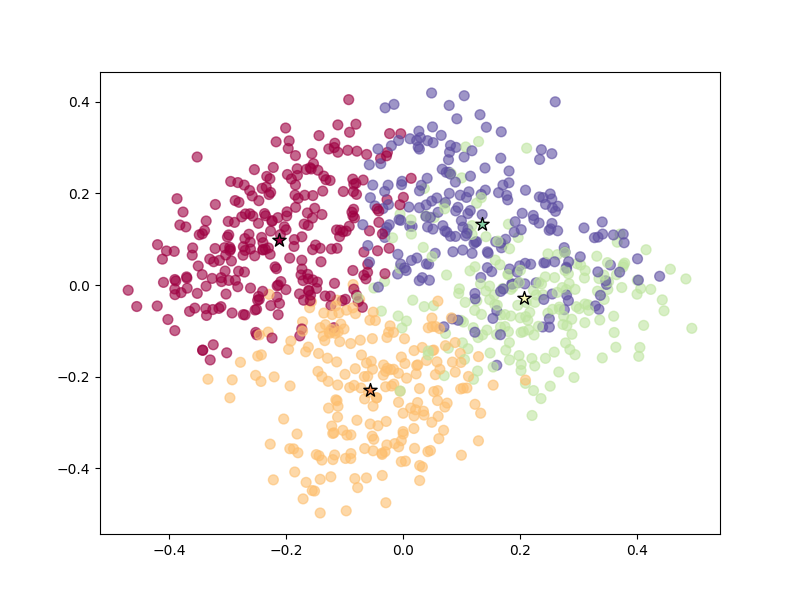}
\caption{\data{SST2}}
\label{fig:imdb_visualization}

  \end{subfigure}

  \caption{Visualization of clustering in \data{AG News} and \data{SST2}, where stars denote cluster centroids.}
  \label{fig:clustering_visualization}
\end{figure}

Figure~\ref{fig:clustering_visualization} visualizes the clustering of sentence embeddings from \data{AG News}, and \data{SST2} datasets, with their dimensions reduced to two using PCA. The illustrations suggest that generic patterns already exist, with instances from various clusters contributing to these patterns.

\section{Demonstration Selection by \sys{FitCF}}
\label{app:demonstration_selection}
\begin{table*}[t!]
    \centering
    \renewcommand*{\arraystretch}{1.2}
    \resizebox{\textwidth}{!}{%
        \begin{tabular}{p{0.48\textwidth} p{0.48\textwidth}}
            \toprule
            \textbf{Text} & \textbf{Counterfactual} \\
            \midrule
            Bovina ends two-year wait. Seventh-seeded Russian Elena Bovina won her first title in two years by beating France's Nathalie Dechy 6-2 2-6 7-5 in the final of the \textbf{Pilot Pen tournament}. 
            & Bovina ends two-year wait. Seventh-seeded Russian Elena Bovina won her first title in two years by beating France's Nathalie Dechy 6-2 2-6 7-5 in the final of the \colorbox{red}{International Event}. \\
            \hline
            \textbf{Wall St.'s Nest Egg} - the Housing Sector  NEW YORK (Reuters) - If there were any doubts that we're  still living in the era of the stay-at-home economy, the rows  of empty seats at the Athens Olympics should help erase them. & \colorbox{red}{The Olympics} - the Housing Sector  NEW YORK (Reuters) - If there were any doubts that we're  still living in the era of the stay-at-home economy, the rows  of empty seats at the Athens Olympics should help erase them.\\
            \hline

            French Take Gold, Bronze in Single \textbf{Kayak} ATHENS, Greece - Winning on whitewater runs in the family for Frenchman Benoit Peschier, though an Olympic gold is something new. Peschier paddled his one-man kayak aggressively but penalty free in both his semifinal and final runs on the manmade \textbf{Olympic} ... & French Take Gold, Bronze in Single \colorbox{red}{Kayaking Competition} ATHENS, Greece - Winning on whitewater runs in the family for Frenchman Benoit Peschier, though an Olympic gold is something new. Peschier paddled his one-man kayak aggressively but without penalty in both his semifinal and final runs on the man-made Olympic \colorbox{red}{course}. \\ \hline

            \textbf{Japanese Utility Plans IPO in October (AP) AP -} Electric Power Development Co., a former state-run utility,\textbf{ said Friday it} is planning an initial public offering on the Tokyo Stock Exchange in October, a deal that could be the country's biggest new stock listing in six years. & Electric Power Development Co., a former state-run utility, is planning an initial public offering on the Tokyo Stock Exchange in October, a deal that could be the country's biggest new stock listing in six years.\\ \hline

            Afghan women make brief \textbf{Olympic} debut Afghan women \textbf{made a short-lived} debut in the Olympic Games on Wednesday as 18-year-old \textbf{judo wildcard} Friba Razayee was defeated after 45 seconds of her first match in the under-70kg middleweight.  & Afghan women make brief debut in \colorbox{red}{international relations} as 18-year-old Friba Razayee was defeated after 45 seconds of her first match in the under-70kg middleweight. \\
            \bottomrule
        \end{tabular}
    }
    \caption{
       The most similar demonstrations selected from each cluster for the question \textit{``Rivals Try to Turn Tables on Charles Schwab By MICHAEL LIEDTKE     SAN FRANCISCO (AP) -- With its low prices and iconoclastic attitude, discount stock broker Charles Schwab Corp. (SCH) represented an annoying stone in Wall Street's wing-tipped shoes for decades...''} from \data{AG News}. Corresponding counterfactuals are generated by \lm{Qwen2.5-72B} using \sys{ZeroCF}.  Differences are marked in \textbf{bold} and edits are highlighted in \colorbox{red}{red}.
    }
    \label{tab:counterfactual_examples}
\end{table*}
Table~\ref{tab:counterfactual_examples} shows the most similar demonstrations selected from each cluster, as shown in Figure~\ref{fig:clustering_visualization} for the question \textit{``Rivals Try to Turn Tables on Charles Schwab By MICHAEL LIEDTKE     SAN FRANCISCO (AP) -- With its low prices and iconoclastic attitude, discount stock broker Charles Schwab Corp. (SCH) represented an annoying stone in Wall Street's wing-tipped shoes for decades...''} from \data{AG News}. 

The decrease in SLFR performance while using a strong LLM can be attributed to the advanced contextual understanding of such models, e.g., \lm{Qwen2.5-72B}. These models are more adept at discerning the underlying context of inputs and therefore less likely to incorrectly flip labels. For instance, as shown in Table~\ref{tab:counterfactual_examples}, the second example remains clearly related to \textbf{business}, as the main topic—Housing Sector—is still evident, even though ``\textit{Wall St.'s Nest Egg}'' is replaced with ``\textit{The Olympic}''.

\section{Dataset}
\label{app:dataset}
\subsection{Label Distribution}
\label{subsec:label_distribution}
\begin{figure}
  \centering

  \begin{subfigure}{\columnwidth}
    \centering
    \centering
\includegraphics[width=\linewidth]{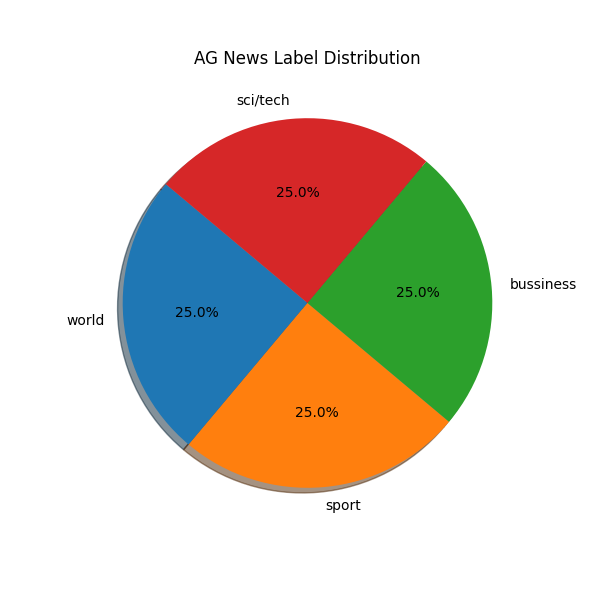}
\caption{\data{AG News}}
\label{fig:ag_news_distribution}
  \end{subfigure}
  \hfill
  \begin{subfigure}{\columnwidth}
\centering
\includegraphics[width=\linewidth]{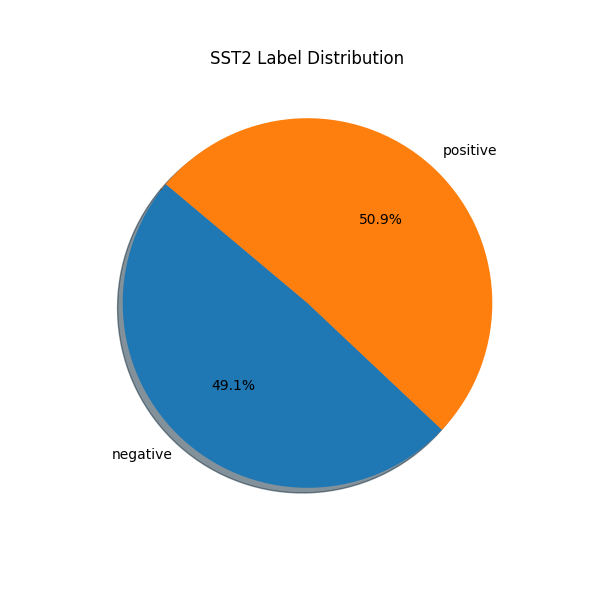}
\caption{\data{SST2}}
\label{fig:imdb_label_distribution}

  \end{subfigure}

  \caption{Label distribution of \data{AG News} and \data{SST2}.}
  \label{fig:label_distribution}
\end{figure}

Figure~\ref{fig:label_distribution} shows the label distributions of \data{AG News} and \data{SST2} validation sets.

\subsection{Dataset Example}
\label{subsec:dataset_example}
\begin{figure}[t!]
\centering
\resizebox{\columnwidth}{!}{
\begin{minipage}{\columnwidth}
\includegraphics[width=\columnwidth]{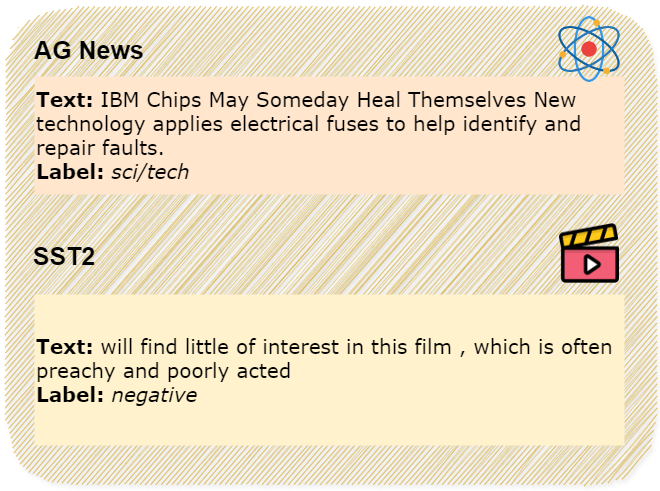}
\end{minipage}
}
\caption{Example instances from \data{AG News} and \data{SST2}.}
\label{fig:dataset_example}
\end{figure}

Figure~\ref{fig:dataset_example} demonstrates example instances and gold labels from \data{AG News} and \data{SST2} datasets.

\section{Experiment}
\label{app:experiment}
\subsection{Models}
Table~\ref{tab:used_model} demonstrates LLMs that are used for \sys{ZeroCF} and \sys{FitCF}. To reduce memory consumption, we use a GPTQ-quantized version \cite{frantar-2023-optq}. All LLMs are directly downloaded from Huggingface and run on a single NVIDIA RTXA6000, A100 or H100 GPU. 

\begin{table*}[t!]
    \centering
    \resizebox{\textwidth}{!}{%
        \begin{tabular}{rccc}

        \toprule
        \textbf{Name}& \textbf{Citation} & \textbf{Size} & \textbf{Link}\\

        \midrule
        \lm{Llama3} & \citet{llama3modelcard} & 8B & \url{https://huggingface.co/meta-llama/Meta-Llama-3-8B}\\
        
        \lm{Qwen2.5} & \citet{qwen2.5} & 32B & \url{https://huggingface.co/Qwen/Qwen2.5-32B-Instruct-GPTQ-Int4} \\
        \lm{Qwen2.5} & \citet{qwen2.5} & 72B & \url{https://huggingface.co/Qwen/Qwen2.5-72B-Instruct-GPTQ-Int4}\\
        
        \bottomrule
        \end{tabular}
        }
    \caption{
    Three open sourced LLMs used in \sys{ZeroCF} and \sys{FitCF}. 
    }
    \label{tab:used_model}
\end{table*}

\subsection{Inference Time}
\begin{table}[t!]
    \centering
    \resizebox{\columnwidth}{!}{%
        \begin{tabular}{c|cc|cc|}

        \toprule
         & \multicolumn{2}{c|}{\textbf{\data{AG News}}} & \multicolumn{2}{c|}{\textbf{\data{SST2}}}\\
         & \sys{ZeroCF} & \sys{FitCF} & \sys{ZeroCF} & \sys{FitCF}\\

        \midrule

        \lm{Llama3-8B} & 8h & 13h & 2h & 5h \\

        \lm{Qwen2.5-32B} & 9h & 17h & 7h & 12h \\

        \lm{Qwen2.5-72B} & 38h & 47h & 8h &  16h \\

        \bottomrule
        \end{tabular}
    }
    \caption{
    Inference time for \sys{ZeroCF} and \sys{FitCF} using \lm{Llama3-8B}, \lm{Qwen2.5-32B} and \lm{Qwen2.5-32B} on \data{AG News} and \data{SST2}.
    }
    \label{tab:inference}
\end{table}
Table~\ref{tab:inference} shows inference time for \sys{ZeroCF} and \sys{FitCF} using \lm{Llama3-8B}, \lm{Qwen2.5-32B} and \lm{Qwen2.5-32B} across \data{AG News} and \data{SST2} datasets.

\section{Calculation of Label Flip Rate}
\label{app:calculation_lfr}
We use the same LLM to serve as both the flip label verifier and the counterfactual generator (\S\ref{subsec:automatic_evaluation}). To validate deployed LLMs' classification performance, we evaluate them on the \data{AG News} and \data{SST2} datasets. Subsequently, we detail the prompt instructions used for flip label verification.

\subsection{Classification Performance of LLMs}

\begin{table*}[t!]
    \centering
    \begin{tabular}{c|ccc}
        \toprule
        \textbf{Dataset} & \textbf{Model} & \textbf{Accuracy} & \textbf{Error Rate}\\

        \midrule
        \multirow{3}{3em}{\data{AG News}} & \lm{Llama3-8B} & \underline{72.39\%} & \underline{0.70\%} \\
        
        & \lm{Qwen2.5-32B} & \textbf{80.73}\% & \textbf{0.28}\% \\
        & \lm{Qwen2.5-72B} & 79.12\% & 0.47\% \\

        \midrule
        \multirow{3}{3em}{\data{SST2}} & \lm{Llama3-8B} & \underline{89.75\%} & 0.00\%\\
        & \lm{Qwen2.5-32B} & \textbf{94.61\%} & \textbf{0.00\%} \\
        & \lm{Qwen2.5-72B} & 94.27\% & \underline{0.11\%} \\
        \bottomrule
    \end{tabular}
    \caption{Accuracy score and error rate on \data{AG News} and \data{SST2} datasets across three runs on the validation set using \lm{Llama3-8B}, \lm{Qwen2.5-32B}, and \lm{Qwen2.5-72B} in a \textit{zero-shot} setting. The error rate is calculated by counting the number of instances where the predicted label falls outside the pre-defined label set.}
    \label{tab:performance}
\end{table*}

Table~\ref{tab:performance} displays the accuracy score and error rate on \data{AG News} and \data{SST2} datasets using \lm{Llama3-8B}, \lm{Qwen2.5-32B}, and \lm{Qwen2.5-72B}. Our findings indicate that \lm{Qwen2.5-32B} demonstrates the best classification performance with the lowest error rate, whereas \lm{Llama3-8B} has the poorest classification performance. Notably, \lm{Qwen2.5-72B} is the only LLM that generates predictions outside the predefined labels on \data{SST2}. Moreover, as selected LLMs and \lm{BERT} perform similarly on the two datasets (Table~\ref{tab:bert_model}), we can assume that the tendency of SLFR will be consistent, and our conclusion should remain unchanged.

\subsection{Prompt Instruction}
\label{app:prompt_lfr}
\begin{verbatim}
You are an excellent assistant for text 
classification. You are provided with an 
original and an edited instance from the 
{dataset_name} dataset. Each instance
belongs to one of {len(labels)} categories: 
{', '.join(labels)}. Determine if the 
predicted classifications of the original 
and edited instances are different.
[original instance] '{instance}'
[edited instance] '{counterfactual}'
Respond with 'yes' if they are different. 
Response with 'no' if they are the same. 
Answer 'yes' or 'no' only!
\end{verbatim}

\end{document}